\def\year{2019}\relax
\documentclass[letterpaper]{article} 

{\noindent\ignorespaces\color{#1}\textbf{#2:}}%
{\par\noindent\ignorespacesafterend}

\newcommand{\eg}{\emph{e.g.,}}
\newcommand{\ie}{\emph{i.e.,}}

\newcommand{\etal}{\emph{et. al}}
\usepackage{aaai19}  
\usepackage{times}  
\usepackage{helvet}  
\usepackage{courier}  
\usepackage{url}  
\usepackage{graphicx}  
\usepackage{times}
\usepackage{epsfig}
\usepackage{amsmath}
\usepackage{amssymb}

\usepackage{gensymb}
\usepackage{balance}
\usepackage{algorithm}
\usepackage[noend]{algpseudocode}
\usepackage{placeins}

\usepackage{siunitx}
\usepackage{tabu}
\usepackage{tabularx}
\usepackage{tablefootnote}
\usepackage{threeparttable}
\usepackage{mathtools}
\usepackage{epstopdf}
\usepackage{float}
\usepackage[T1]{fontenc}
\usepackage{enumitem}
\usepackage[percent]{overpic} 
\usepackage{color}

\graphicspath{ {plots/} {diagrams/} {figures/}}

\begin{document}

\title{Learning to Learn in Simulation}
\author{Ervin Teng \ and Bob Iannucci\\
Carnegie Mellon University\\
Moffett Field, CA \\
\{ervin.teng, bob\}@sv.cmu.edu \\
} 





\maketitle

\begin{abstract} 

Deep learning often requires the manual collection and annotation of a training set. On robotic platforms, can we partially automate this task by training the robot to be curious, \ie{} to seek out beneficial training information in the environment? In this work, we address the problem of curiosity as it relates to online, real-time, human-in-the-loop training of an object detection algorithm onboard a drone, where motion is constrained to two dimensions. We use a 3D simulation environment and deep reinforcement learning to train a curiosity agent to, in turn, train the object detection model. This agent could have one of two conflicting objectives: train \textit{as quickly as possible}, or train \textit{with minimal human input}. We outline a reward function that allows the curiosity agent to learn either of these objectives, while taking into account some of the physical characteristics of the drone platform on which it is meant to run. In addition, We show that we can weigh the importance of achieving these objectives by adjusting a parameter in the reward function. 


\end{abstract}




\section{Introduction}
\label{sec:intro}
Today's deep learning algorithms are extremely effective at many computer vision tasks, ranging from object detection to image segmentation, \textit{once they have been trained}. But what if we want to retrain one quickly? Consider this scenario: a drone is equipped with an object detector and is on a mission to detect and track a particular subject individual. The individual steps into a vehicle. A human operator would have no problem refocusing on the vehicle. But our drone would need to be re-directed and re-trained to detect that particular vehicle. Because the mission is unfolding in real-time, a stop-and-retrain approach (\eg{} gathering new images of the target car, building a new training set, re-training the object detector, and re-deploying the detector to the drone) would be ineffective. We can imagine some means for in-flight re-training with input from a human user. 

But in doing so, have we just added the burden of training to the already-busy user? As much as possible, we would like the autonomous agent---in this case the drone---be able to learn independently. To tackle this problem, we use the analogy of an actual student. A good student both \textit{studies well} and \textit{is curious}---\ie{} they extract as much knowledge as they can from the information they have, and seek new information as needed. They may have guidance from a teacher or mentor, but this guidance is limited and should be used sparingly. Similarly, we can imagine a system that trains the drone's object detector by both extracting information from the drone's video feed, moving it around to get more views, and asking the human user for ground truth. 

\begin{figure}[t] \centering
    \includegraphics[width=\linewidth]{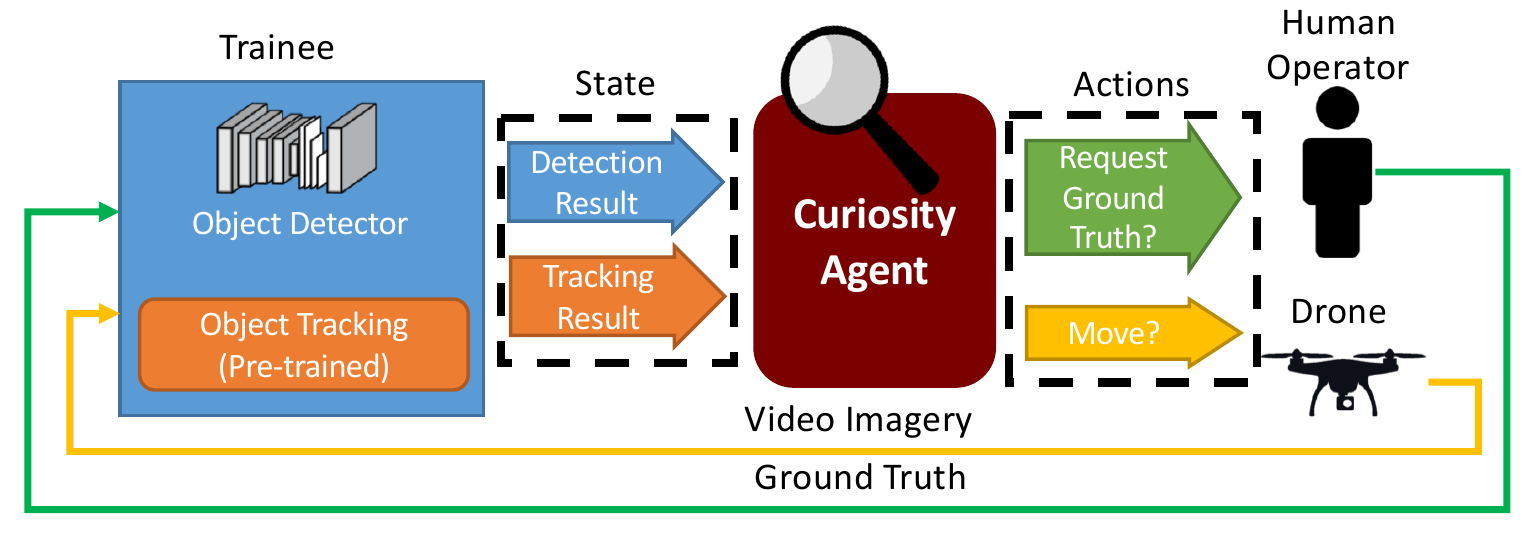}
    \caption{Diagram of autonomous curiosity can assist real-time online training of an object detector onboard a robotic platform (\eg{} a drone). The curiosity agent receives feedback from the trainee, which consists of the object detector being trained and an object tracker which produces training events as long as the target is tracked. It is responsible for moving the drone to get a better view, if necessary, or asking the human user for ground truth~\cite{Teng2019AutonomousAgents}}
    \label{fig:systemdiag}
\end{figure}

In our prior work~\cite{Teng2019AutonomousAgents}, we introduced autonomous curiosity, a way for a pre-trained curiosity agent to move a drone around to increase how beneficial (the \textit{training benefit}) each ground truth user interaction is in training the drone's model. Figure~\ref{fig:systemdiag} shows how our system interacts with both the drone's navigational system as well as the human user. We showed that for constrained motion, a trained \textit{curiosity agent} can outperform na\"{i}ve approaches in terms of increasing the training benefit of each user interaction required to train an object detector. As learning to learn in the real world would take a prohibitively long amount of time, we trained our curiosity agent in a simulation created using the Unity game engine and ML-Agents API~\cite{Juliani2018Unity:Agents}.

In this paper, we expand the problem setup of autonomous curiosity to two dimensions and four movement actions---towards and away from the subject in addition to orbital movement. This introduces an additional consideration---we cannot make the assumption that all motions of the drone take the same amount of time~\cite{Teng2019AutonomousAgents}. We hypothesize that agents trained without a notion of the time-cost of each of its actions, on the specific robotic platform they will be deployed, will not perform as well as one which was trained with this in mind. This paper makes the following contributions towards solving this problem.
\begin{itemize}[noitemsep]
\item A training environment that enables two-dimensional agent training, and a fused deep-Q model curiosity agent architecture that can account for these degrees of freedom. 
\item A reward structure that enables the agent to learn the variation in time between different types of actions. 
\item A validation of this reward structure, showing that we can shape the behavior of the agent to both optimize for training benefit per unit time as well as training benefit per user interaction.
\end{itemize}

This paper is structured as follows. Section~\ref{sec:background} gives an overview of the problem domain and related work. Section~\ref{sec:problem} formulates the training of an object detector onboard a drone platform for the 2-dimensional movement space. Section~\ref{sec:architecture} describes the architecture of the curiosity agent itself and the reward structure. Finally, Section~\ref{sec:experiments_simulation} describe the experiments that we used to validate the design of our reward function.

\section{Background and Prior Work}
\label{sec:background}

\subsection{Problem Domain}
\label{sec:toot}

For mid-mission re-training of deep learning models, the video frames and corresponding annotations generated by the human user become the training set. With an object detector that has a fixed sample efficiency, the remaining figure-of-merit is the improvement of the model at each step. 

In prior work~\cite{Teng2019AutonomousAgents}, we primarily examined the improvement garnered from a single user interaction. We define a training \textit{episode} as a single training session with one target object. If we divide the episode into discrete time-steps, at each time-step $i$, the user can provide ground truth to the system. This forms a vector $U$ of user interactions:
\begin{equation}
U = (u_1, u_2, ..., u_n),   u_i = \begin{cases} 
      1 & \text{user int. on frame $v_i$} \\
      0 & \text{otherwise}
   \end{cases}
\end{equation}
The metric in question, then, is the average \textit{incremental training benefit}~\cite{Teng2017,Teng2018ClickBAIT-v2:Real-Time} of each user interaction during an episode. Qualitatively, incremental training benefit is a measure of the increase in model performance (average precision in this case) that can be attributed to a particular user interaction. For a fixed number of time steps within an episode, we can compute the average ITB from time step $i$ to $j$ with:

\begin{equation}
\overline{ITB}_{u_{(i,j)}} = \frac{P_j - P_i}{\sum_{x=i}^{j} u_x} 
\label{eqn:meanitb}
\end{equation}
where $P_i$ is the performance (\ie{} the AP) of the trainee model at step $i$, and $P_0$ is reserved for the performance of the object detection model before any online training~\cite{Teng2019AutonomousAgents}. In this paper, we will use the final $\overline{ITB}_{u_{(1,f)}}$ of all $f$ user interactions in the episode as a measure of the effectiveness of the user interactions.

In real-world operation, we are also concerned with the time-efficiency of our training procedure. Therefore, we also consider the $\overline{ITB}$ per unit elapsed time:
\begin{equation}
\label{eqn:ITBtvid}
\overline{ITB}_{t_{elapsed}} = \frac{P_j - P_i}{t_{elapsed{(i,j)}}}
\end{equation}
where $t_{elapsed{(i,j)}}$ is the real time elapsed between frame $v_i$ and frame $v_j$.
In this work, we will examine both $\overline{ITB}_{t_{elapsed}}$ and $\overline{ITB}_{u_i}$ as optimization goals for our curiosity agents.

\subsection{Related Work}


\textbf{Active learning} starts with the premise that ground truth annotations for training data comes from a human ``oracle'', and that this human's time is limited and valuable. Active learning attempts to select a subset of the available training set which would produce the best trained model. A number of heuristics, including expected model change, expected loss reduction, or core-set coverage, are often used to determine the value of each annotation request~\cite{Settles2010,Yao2012InteractiveDetection,Sener2017}. Reinforcement learning agents~\cite{Fang2017} can be used for active learning in streaming applications where the whole training set is not known \emph{a priori}. In effect, this agent \textit{learns} what constitutes a valuable datapoint, and can predict the usefulness of new incoming data.


\textbf{Active perception}, in contrast to active learning, introduces an element of spatial awareness. While active learning addresses \textit{what} to look at, active perception addresses \textit{where} to look. Often, this means guiding visual search within one image for object localization and classification, performing operations only on a subset of the image at a time. This has significant computational advantages for high-resolution images, and can be achieved through supervised learning~\cite{Ranzato2014} or reinforcement learning~\cite{Mathe2016,Caicedo2015}. Reinforcement learning has also shown promise in being able to explore and find activities across time as well (\ie{}, in videos)~\cite{Yeung2016End-to-endVideos}. In cyber-physical systems, such as robots, active perception can be used for real-world actuation~\cite{Jayaraman2016Look-aheadMotion,Malmir2015}. As in active learning, reinforcement learning has shown success in training agents \textit{how to learn}, in this case how to explore a large image in such a way that the imagery gathered enables another part of the model to learn about the image as a whole~\cite{Jayaraman2018}.

Both fields seek to solve the problem of \textit{information acquisition}, whether from a human or from the environment. In this work and in our prior work~\cite{Teng2019AutonomousAgents}, we combine active learning and active perception functionality into a single reinforcement learning agent.

\section{Problem Setup}
\label{sec:problem}
We allow the drone to move in two dimensions above the ground. Recognizing that changes in altitude may not be desirable from an operational and safety point of view, we focus on allowing the drone to move freely within a planar disk centered above the subject. Onboard the drone, there are two machine learning-based components. The object detector \textit{trainee} is trained online by annotations and imagery it receives from a pre-trained \textit{curiosity agent}. In this section, we describe the environment and trainee as forming the problem or \textit{game} that the curiosity agent needs to solve.

We will refer to a generated scenario with a fixed subject and some obstructions as a \textit{scene}, an image of the subject from a point on the orbit as a \textit{view}, and the set of all possible views of the subject (one for each point in the orbit) as the \textit{exploration space}~\cite{Teng2019AutonomousAgents}.

\subsection{Environment}

\begin{figure}[t] \centering
    \includegraphics[width=0.7\linewidth]{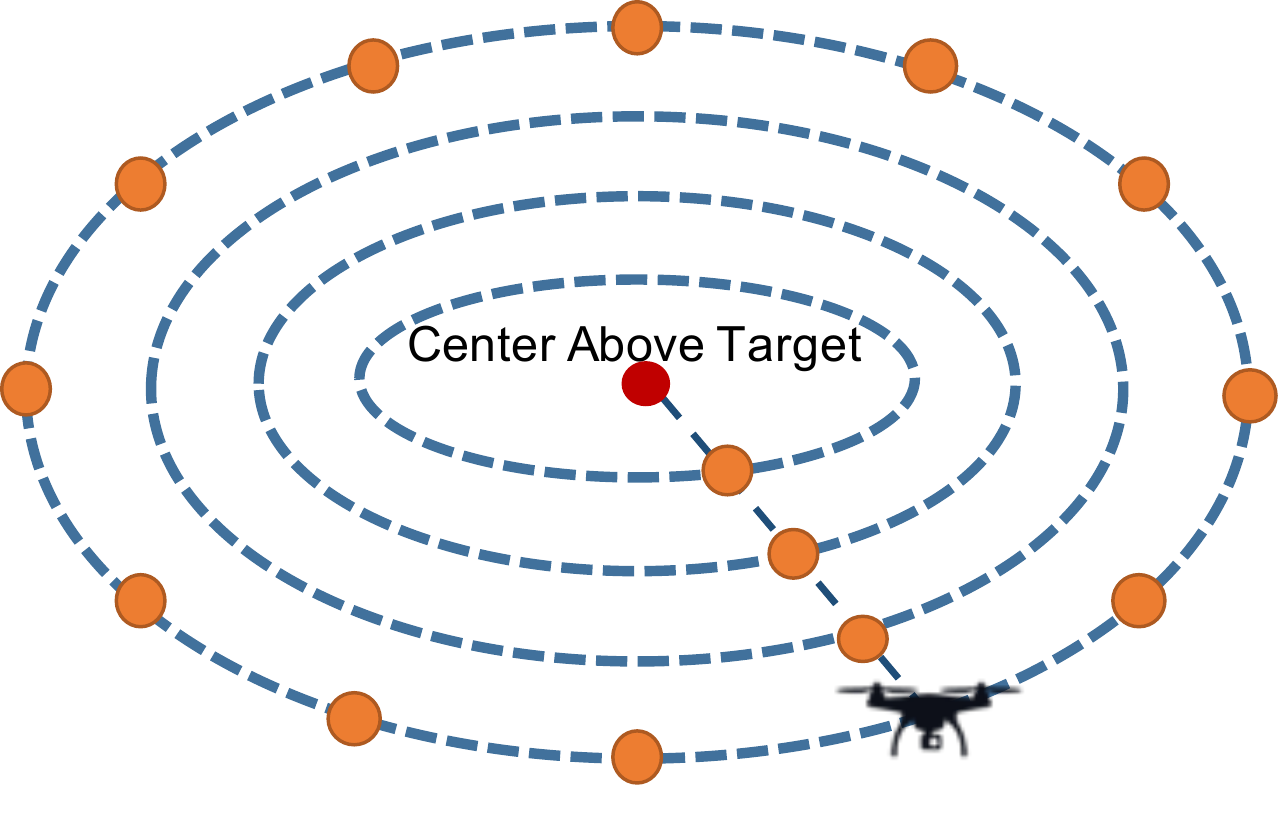}
    \caption{Discretized planar disk exploration space. From a given point (orange dot) in the exploration space, the drone can move towards or away from the center of the disk along the radius between it and the center, or around the center along the orbit equidistant from the center. One way to describe this exploration space is a series of concentric orbits on which points exist at discrete angular intervals.}
    \label{fig:concentric_orbits}
\end{figure}

To make the movement primitives subject-centric, movements are either an angular step around an orbit or a radial move towards or away from the center of the circular disk. The outer orbit of the disk is positioned relative to the subject with respect to two parameters, the elevation angle $\theta$ and the line-of-sight distance $d$ from the subject to the drone. In our orbits, we set $\theta = 30\degree$, and $d$ is variable depending on the subject. Figure~\ref{fig:concentric_orbits} shows how this expanded degree of freedom affects the exploration space.

Each point along the orbit path is separated by a fixed angle $\Delta \alpha$. We chose this angle to be $\Delta \alpha = 12 \degree$---roughly 1/3 of the $37.2\degree$ FOV of a GoPro Hero 4 Black in Narrow mode, cropped to a square (the imaging setup used in our field experiments)~\cite{Teng2019AutonomousAgents}. Furthermore, each radial inwards towards the subject is divided up into evenly-spaced segments. This type of movement could also be described as a series of concentric, evenly-spaced orbits, with the smallest of radius $r_{min}$, the largest of radius $r_{max}$, and $\Delta r$ meters between subsequently sized orbits.

\begin{figure}[t] \centering
    \includegraphics[width=0.6\linewidth]{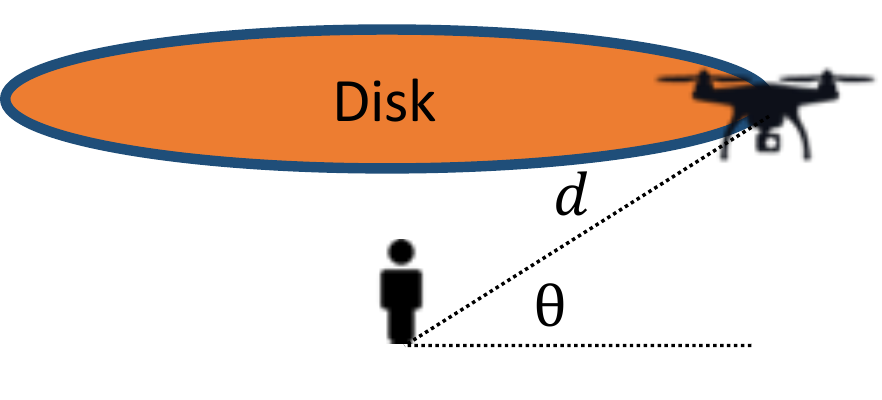}
    \caption{Discretized planar disk exploration space, as viewed from the side. The person in the center represents the subject. $d$ and $\theta$ define the position of the disk relative to the subject.}
    \label{fig:disk_above}
\end{figure}

This entire planar disk is centered above and positioned as shown in Figure~\ref{fig:disk_above}, described by $\theta$ and $d$.

\subsection{Simulation}

Learning to learn through reinforcement learning, as we are doing here, requires thousands of iterations on different scenes. Doing so with real flights would be incredibly difficult, if not impossible. How do we train the curiosity agent in a virtual environment? Sadeghi \etal~\cite{Sadeghi2016CAD2RL:Image} found that realism is less important than variety for generalizing algorithms from simulation into the real world. In our prior work~\cite{Teng2019AutonomousAgents}, we use this principle to develop a simple scene simulator that creates many differently colored scenes, and show that the behavior of an agent trained in this environment is similar to the same agent evaluated on real imagery. We use an extended version of that simulation environment here. 

We procedurally generate scenes using the Unity 3D game engine, which interacts with the Tensorflow DQN network through the ML-Agents~\cite{Juliani2018Unity:Agents} API. At the beginning of each episode, we create a scene by first choosing one of 7 different subjects. The additional two subjects (train and ship) were chosen to be long and awkwardly shaped, making it both difficult for the object tracker, and completely different in shape and appearance from different views. This was in part to emulate the difficult helicopter scenario from~\cite{Teng2019AutonomousAgents}. The subject is placed in the center of our scene, and assigned a random hue. We perturb the center point of the orbit by up to 5m (conservative estimate of GPS accuracy) to simulate inaccuracy in aiming the camera.

A number of obstructions $n_{obs}$, from eight different types of randomly assigned color, are then placed on the ground in a region around the subject within the diameter of the orbit. To add variety to the scenes, we also randomly rotate and stretch the obstructions, while ensuring that the height does not exceed that of the flight disk. If an obstruction is detected to be overlapping the subject, it is not placed. For a standoff distance $d = 60$ (simulated) meters, we set the inner radius of the placement region to 10m and the outer to 35m.

One of the main differences between simulated scenes and real scenes is the presence of similar objects to the subject (\ie{} additional cars and people in the car and person scenes). This means that the trainee model must also learn to differentiate objects of the same class, but, for instance, in different colors. To emulate this, we also add, with the same parameters as the obstructions, two additional duplicate subjects--confusors--of random hue. We ensure these confusors are of a different hue than the actual subject.


During training, we set $\Delta \alpha = 12\degree$, and create 6 concentric orbits. For a standoff distance $d = 60$ meters, this means each orbit is $\delta r = 8.66$ meters from the next.

\begin{figure} \centering
    \includegraphics[width=\linewidth]{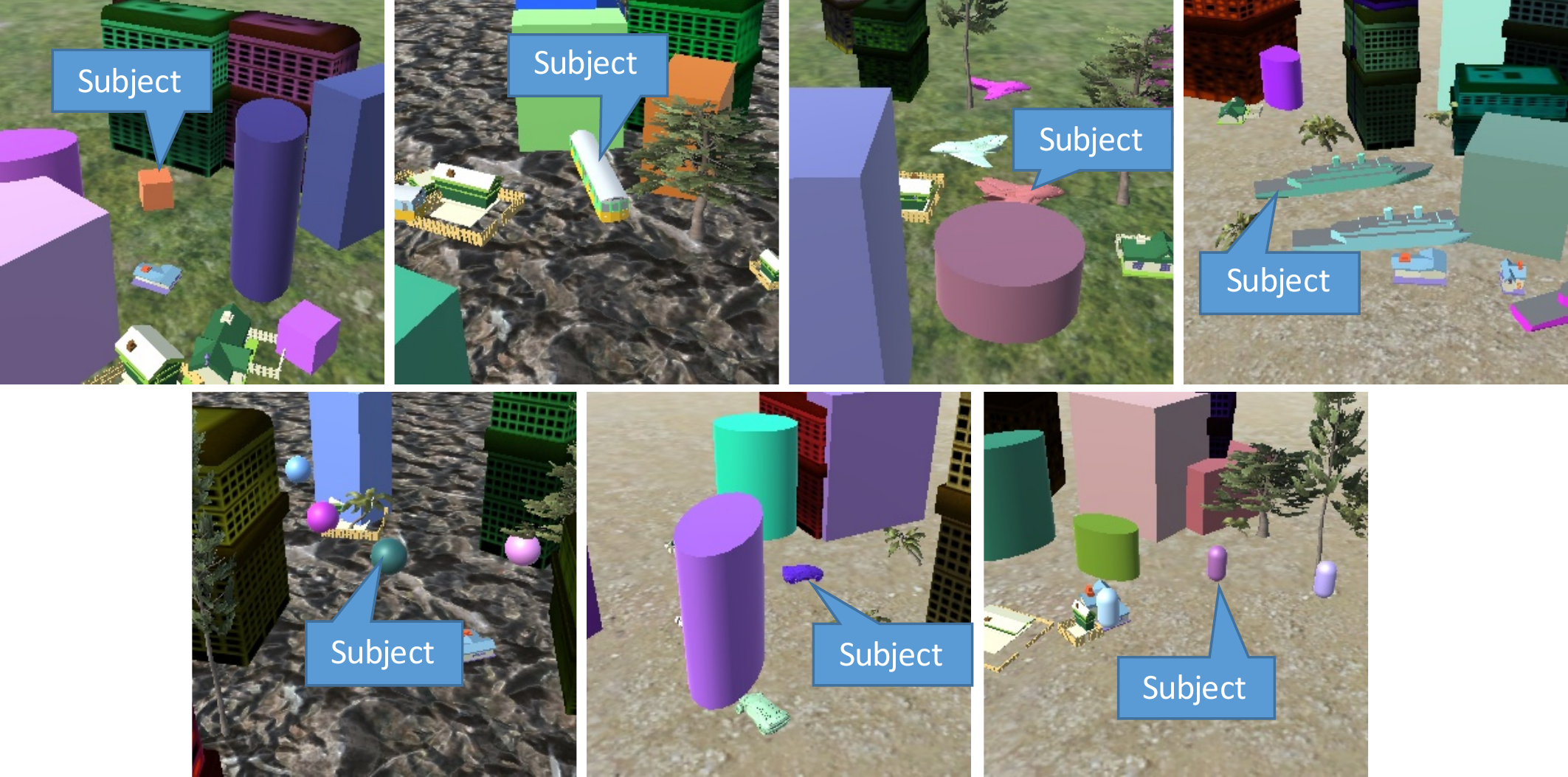}
    \caption{Views of generated scenes for the 7 types of subjects From left to right, top to bottom: cube, train, plane, ship, sphere, car, capsule. Note the duplicate subjects of various colors int the scenes.}
    \label{fig:curiosity2d_simscreenshots}
\end{figure}

\subsection{Trainee}

The trainee is based on a system we developed for real-time training of an object detector~\cite{Teng2018ClickBAIT-v2:Real-Time}. It consists of two main parts: an object tracker~\cite{Valmadre2017} that assists in gathering training data, and an object detector~\cite{Liu2016} that is being fine-tuned in real-time. If ground truth is given by the user in the form of a bounding box, the object tracker is initialized. As long as the tracker is sufficiently confident about its tracking result, each tracked frame is used to initiate one minibatch training round (one gradient descent update). This continues until tracking fails, at which time the system is idle until the human user again gives ground truth. This system allows for training to continue without ground truth on every single frame.

\subsection{Actions}

The curiosity agent chooses actions that would navigate the drone within the exploration space, or request ground truth annotations from the human operator. Actions taken by the curiosity agent should influence the drone's position within the exploration space, the state of the trainee, or both. The curiosity agent can take advantage of the trainee's object tracking abilities to avoid asking for ground truth at every point along the orbit.

At each action-step $i$, the agent can take one of four actions from the set of actions $A=\{0,1,2,3,4,5\}$:
\begin{itemize}[noitemsep]
\item Don't move. If object tracking is initialized, perform a training round.
\item Request user interaction, which produces a ground truth bounding box. Ground truth is then used for a training round and to initialize object tracking.
\item Move left. The drone moves left one spot in its current orbit. If object tracking was successful prior to the move, we train on the previous view during the transit.
\item Move right. The drone moves right one spot in its current orbit. If object tracking was successful prior to the move, we train on the previous view during the transit.
\item Move forwards. The drone moves to the next innermost orbit, if available. If object tracking was successful prior to the move, train on the previous view during the transit.
\item Move backwards. The drone moves to the next outermost orbit, if available. If object tracking was successful prior to the move, train on the previous view during the transit.
\end{itemize}

During an episode, the curiosity agent can interact with the user by requesting user interaction, where the user is sent the current drone view of the subject and receives a bounding box ground truth from the user. In addition, if ground truth is not received, the current tracked bounding box, if present, is used to initiate a training round for the trainee. The full process can be seen in the Appendix, Algorithm~\ref{alg:curiosityagent2}.


\subsection{Timing and Time-Model}
\label{sec:curiosity2d_timing}

\subsubsection{Time Progression}
Because movement actions no longer take the same amount of time, the assumption that all actions take a fixed amount of time is unrealistic. Rather, we must clarify the previously simple notion of a time-step, \ie{} a period of time in which the agent chooses and executes an action. To avoid confusion, a step in which the agent chooses an action will be referred to as an \textit{action-step}, which may correspond to varying amounts of actual time passing.

Each action at action-step $i$ results in an increment of time $t_i$. The time elapsed for a particular action can be modeled by the following parameters:

\begin{itemize}[noitemsep]
\item The time it takes for a user annotation $t_{click}$
\item The time required for a trainingRound command $t_{train}$
\item The airspeed in meters per second of the drone $s_a$, assumed constant
\item The distance in meters of a radial movement step $\Delta r$
\item The size in degrees of an orbital movement step $\Delta \alpha$
\end{itemize}

A movement step, \ie{} $a_i = 2,3,4$ or 5, would at minimum incur a time cost $t_{move}$ of $2\pi r \frac{\Delta \alpha}{360\degree}/s_a$ or $\Delta r/s_a$, depending on the direction of the motion. We also assume that at each action-step, the drone can simultaneously move while training. If a training round takes place, the complete movement step will take $t_i = \max{(t_{move},t_{train})}$. The actual values of these parameters are dictated by the physical characteristics of the drone the agent is operating, and the compute platform on which the trainee is being executed.

A requestGroundTruth event ($a_i=0$) takes a fixed amount of time , but if it results in a trainingRound, that time becomes $t_i = t_{train} + t_{click}$. If it does not, this becomes $t_i = t_{click}$. 


\subsubsection{Episode Termination}

Minimizing elapsed time combined with a fixed episode length of action-steps, however, would introduce a conflicting objective: would the agent choose to minimize the time elapsed per action-step, or maximize the positive reward (gain in performance) of the trainee model? We can remove this conflict by turning one of these two goals into a binary, achievable end condition rather than an optimization objective. A sufficiently high reward for achieving this so-called ``win state'' means that the trained agent will \textit{always} try to achieve this goal, then optimize the other objective as a secondary goal.

We create a win condition using the gain in performance of the trainee model, with the intuition that in real-world use, we will always want to train the trainee model to a good performance; doing it quickly is an added bonus. Furthermore, an agent that is able to increase the performance of a trainee model quickly is likely to be able to continue improving it if the system is not terminated after the win condition is reached. Within the condition of reaching the win state, the agent can then attempt to reduce the needed number and duration of action-steps.

An effective win condition is sufficiently challenging so as to not be trivial, but also achievable for all scenes. A win condition that is rarely reached, or is impossible to reach for certain scenes, will lead the agent to never learn to reach it. We ensure this by adding two independent conditions. First, we consider it a win if the trainee model reaches an average precision (AP) of 0.85. However, for very difficult scenes, \eg{} if the object is small or irregular,  where AP begins around 0 and will take an unreasonably long time to reach 0.85, we add a second condition: if the AP increases by 0.70, a win is declared. Both these objectives ensure that a winning agent is effective in training the trainee model, without being impossible.

\section{Approach}
\label{sec:architecture}
This section describes the design of the reward function, the representation of state for the agent, and the architecture of the agent itself. In particular, we will focus on what is needed to incorporate both the expanded range of motion and an explicit notion of elapsed real time. 

\subsection{Reward Function}

\subsubsection{Intermediate Rewards}

Our reward function has two components, a positive reward for improving the trainee model, and a negative reward that should be minimized. The final reward, $\mathcal{R}_{T}$, was described with a weighted sum between these two functions. 

\begin{equation}
\mathcal{R}_{T} = w_p\mathcal{R}_{l} + (1-w_p)\mathcal{R}_{n}
\label{eqn:curiosity_2d_totalreward}
\end{equation}
where $w_p$ is the weight of the positive learning reward, and can be regarded as an application-specific design parameter. The negative reward encompasses both behavior that needs to be minimized as well as the minimization of time. We split the negative reward into two components:
\begin{equation}
\mathcal{R}_{n} = w_n\mathcal{R}_{behave} + (1-w_n)\mathcal{R}_{time}
\label{eqn:curiosity_2d_negreward}
\end{equation}
where $\mathcal{R}_{behave}$ and $\mathcal{R}_{time}$ are the negative rewards associated with behavior and elapsed time, respectively. $w_n$ is the weighing between these two components. Both $w_n$ and $w_p$ are defined such that the magnitude of the total reward $\mathcal{R}_{T}$ does not exceed 1. 

\textbf{Positive Learning Reward:} As in the prior set of experiments, we use the incremental gain in AP for that action-step, \ie{} the \textit{incremental training benefit} (Section~\ref{sec:toot}) of any training event that happened during that action-step. The reward function after action-step $i$, $0\leq i < T$ is given by the capped difference in training objective loss

\begin{equation}
\mathcal{R}_{l} = \min\{\max \{10 (P_i - P_{i-1}),-1\},1\}
\end{equation}

Where $P_i$ is the performance (AP) of the object detector trainee, trained up to action-step $i$. We multiply the difference by a factor of 10 to bring the typical difference in AP, given the hyperparameters of our trainee, closer to 1. 

If the exploration space was a single orbit, it is practical (< 1 second on an NVIDIA GTX 1080) to evaluate the trainee over all views in the full exploration space at each action-step. This is still the ideal behavior. However, as the exploration space is much larger, training could take an impractical amount of time. To address this, during training, the trainee is only evaluated against a random subsampling of the views in the exploration space. This subsample, selected to be 30 views so that the evaluation time is approximately 1 second, is selected only at the beginning of each episode to reduce the noise in the reward signal throughout the episode. 

\textbf{Negative Behavioral Reward:} We maintain the ability to negatively punish behavior that should be used sparingly, such as asking the user for input. Therefore, when $a_i = 1$, 
\begin{equation}
\mathcal{R}_{behave} = -1
\end{equation}

\textbf{Negative Time Reward:} To force the agent to minimize time elapsed, we assign a small negative reward at each action-step. To capture the notion that each action-step consumes variable real time, we assign the negative reward based on the time $t_i$ elapsed at each action step (Section~\ref{sec:curiosity2d_timing}). 
\begin{equation}
\mathcal{R}_{time} = -c_t t_i
\end{equation}
where $c_t = 0.08$ is a weighting parameter so that the magnitude of $\mathcal{R}_{time}$ is relatively small (sum of about 1 for the entire episode) compared to the other rewards. This is important, as unlike $\mathcal{R}_{behave}$, $\mathcal{R}_{time}$ is cumulative throughout the episode and is never zero. $c_t = 0.08$ was chosen so that $\mathcal{R}_{time}$ = -0.2 if the action-step $i$ takes 2.5 seconds.

\subsubsection{Final Rewards}

In Section~\ref{sec:curiosity2d_timing}, we describe when an episode is terminated due to a win condition. If an episode terminates with no win, \ie{} times-out after 10,000 action-steps, $\mathcal{R}_{T}$ is assigned as normal for the final action-step. However, if the episode terminates because one of the two win conditions was reached, we assign $\mathcal{R}_{T} = 1$. This is substantially larger than typically achieved via the positive learning reward, making the episodes where a win happened much more valuable than others. 

\subsection{State Representation}

An effective curiosity agent should, at minimum, have a notion of (a) the view at its current location, (b) the current output of the trainee, and finally (c) the relative position within the exploration space. Here, we adapt this principle to include the expanded degree-of-freedom. 

\textbf{Positional Information: }We have two positional variables required to describe the drone's position on the disk to the agent. This constitutes a positional vector $P$, where $P = [p0, p1]$, reserving the subscript to describe action-step. $p0$ is the angular positional scalar, with $p0_i = \alpha / 360$, where $\alpha$ is the heading angle from the target to the drone on the orbit. $p1$ represents the radial positional scalar, $p1_i = r / r_{max}$, where $r$ the radial distance from the drone's current position to the center of the exploration space disk. 

\begin{figure} \centering
    \includegraphics[width=0.9\linewidth]{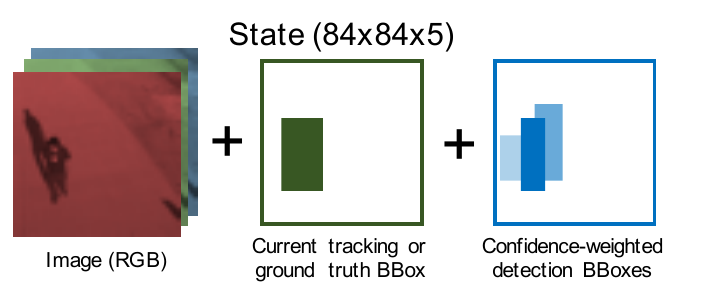}
    \caption{Matrix state representation for Q-Network. The RGB channels of the view are concatenated with representations of the current tracked bounding box and the current output of the trainee object detector, respectively~\cite{Teng2019AutonomousAgents}. The output of the object detector consists of filled boxes whose magnitude is dependent on the confidence of the output.}
    \label{fig:staterep}
\end{figure}

\textbf{Trainee Feedback: } We also pass a matrix state to give the curiosity agent an idea of how the model it is training is doing on the current view. To do this, we concatenate, as an 84x84x5 matrix (Figure~\ref{fig:staterep}): 
\begin{itemize}[noitemsep]
\item The current view $v$.
\item Filled-in bounding box (value $=255$) of the current ground truth BBox (as provided by the user) or tracked BBox, if object tracking is active.
\item Confidence-weighted BBox (value $=255*c$) of the detected bounding boxes, drawn in order of lowest confidence first (higher confidence boxes overlap and cover lower ones). Threshold for SSD output is 0.3.
\end{itemize}

\subsection{Curiosity Agent Architecture}

\begin{figure} \centering
    \includegraphics[width=0.9\linewidth]{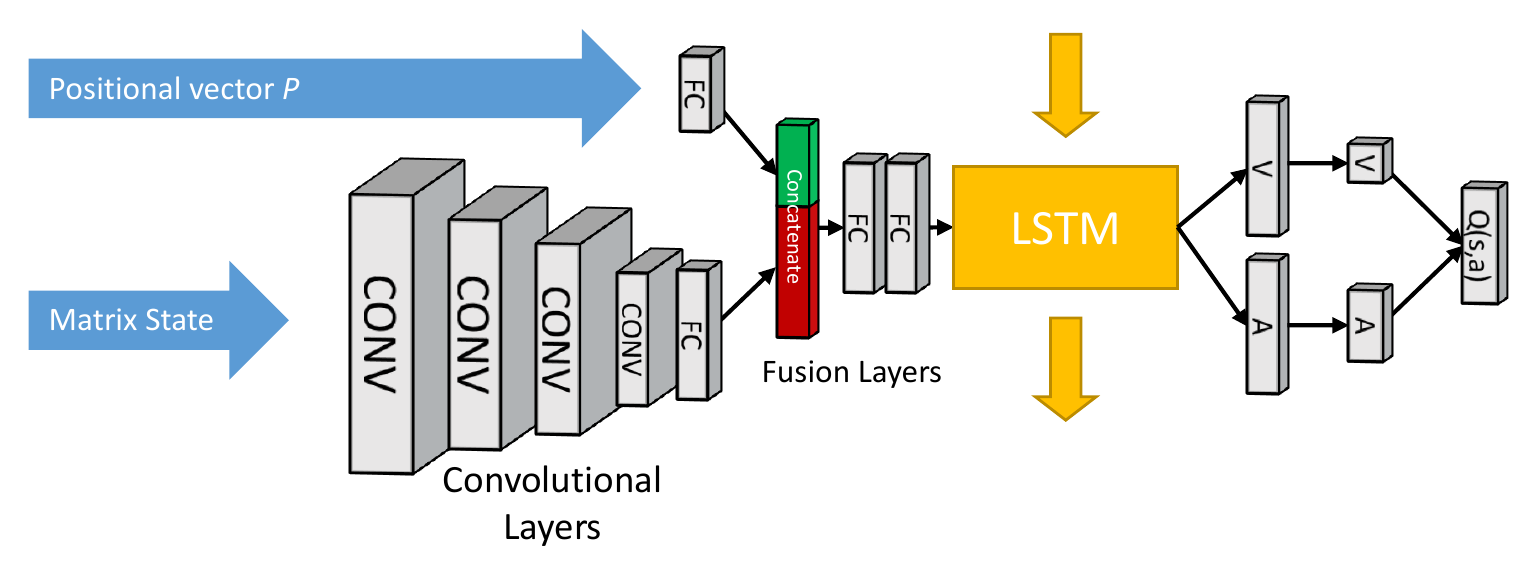}
    \caption{Structure of deep Q network with concatenated state. An LSTM gives temporal notion to the network, and a dueling setup adds additional stability during training. The matrix feature extractor is composed of 4 convolutional layers having respectively, 32, 64, and 64 output channels with filters of sizes 8x8, 4x4, and 3x3, and strides 4, 2, and 1. It is followed by a fully-connected layer with 512 outputs. The positional vector feature extractor is a fully-connected layer with 16 outputs. This is concatenated with the output FC layer of the convolutional network. Two additional fully-connected layers of 512 outputs each are used to fuse the information extracted from both states before the LSTM.}
    \label{fig:DQN-Fused}
\end{figure}

We have a discrete set of six actions and a state that consists of two parts, an 84x84x5 matrix as well as a vector encoding the drone's position within the exploration space. While we maintain the recurrent~\cite{Hausknecht2015DeepMDPs}, dueling~\cite{Wang2016DuelingLearning}, and double-Q~\cite{vanHasselt2016DeepQ-learning}, we revise how information from our state is translated into action. Figure~\ref{fig:DQN-Fused} shows this architecture. For the matrix state, we use convolutional layers as feature-extractors; correspondingly, we use a single fully-connected layer to do the same for the positional vector $P$. These outputs are concatenated and passed through two additional fully-connected layers, that serve to fuse the information found in both parts of the state. This allows the agent to make action decisions based on a relationship between both \textit{what it sees} and \textit{where it is}. 

\section{Experiments and Results}
\label{sec:experiments_simulation}
We hypothesized that an agent trained without a notion of the physical characteristics of the drone it is controlling would perform suboptimally. To validate this hypothesis, we trained multiple agents using varying values for the training round time $t_{train}$ and the drone airspeed $s_a$. These values correspond to the capabilities of the onboard computing platform, and the desired flight characteristics of the drone itself, respectively. We use a realistic value for $t_{click} = 0.9$ seconds~\cite{Bearman2016WhatsSupervision}. We then compared the $ITB_{t_{elapsed}}$ for these agents, given that they are operating \textit{with} and \textit{without} matching performance characteristics.

To show that we can still train for $\overline{ITB}_{u_{(1,f)}}$, we varied $w_n$. However, we also showed that the choice of $w_n$ does present a tradeoff between $\overline{ITB}_{u_{(1,f)}}$ and $\overline{ITB}_{t_{elapsed}}$.

\subsection{Adaptation to Physical Characteristics}

\subsubsection{Experiment Setup}

To examine how behavior and incremental training benefit over time change with varying physical characteristics, we trained two agents, \textbf{Agent A} and \textbf{Agent B}.
\begin{itemize}[noitemsep]
\item \textbf{Agent A} has a drone with $s_a = 2.5$ meters/second, and a compute platform with $t_{train} = 0.305$ seconds, equivalent to an NVIDIA GTX1080.
\item \textbf{Agent B} has a drone with $s_a = 10.0$ meters/second, and a compute platform with $t_{train} = 2.5$ seconds, equivalent to an NVIDIA Jetson TX2.
\end{itemize}

Both of these agents were trained with a simulation environment that matched their respective characteristics. They were then evaluated on the same 50 generated scenes, each with a random $n_{obs}$ between 20 and 25. In the evaluation scenes, the episode is terminated by elapsed time, not the AP reaching the appropriate threshold---as it would be in the real-world, when the drone' battery terminates. When elapsed time reaches 5 minutes, we terminate the episode and measure the $\overline{ITB}_{t_{elapsed}}$ during that sequence.

Each agent was evaluated on scenes simulating the physical characteristics of their own drone platform, and that of the other agent. For instance, Agent A was used on a virtual drone that matched its $s_a$ and $t_{train}$ values, as well as one that matched Agent B's physical characteristics. This experiment was intended to show the effect of training towards a particular platform. 

\subsubsection{Results}


 At each action-step, we evaluate the trainee performance (AP) over all of the views in the scene. We would like to plot model performance $P_i$ with the independent variable being \textit{elapsed time} $t_{elapsed}$, not action-steps $i$. However, this means we cannot simply average all 50 test scenarios, as each action-step may have taken place at a different time-step. Instead, we bin each action-step that occurs into a 1-second bin, and average all the $P$ values from all 50 traces in that bin. We do this for all 300 seconds of the episodes. We refer to this re-indexed performance value as $P_{t_{elapsed}}$.
 
 \begin{figure}[ht] \centering
    \includegraphics[width=1.0\linewidth]{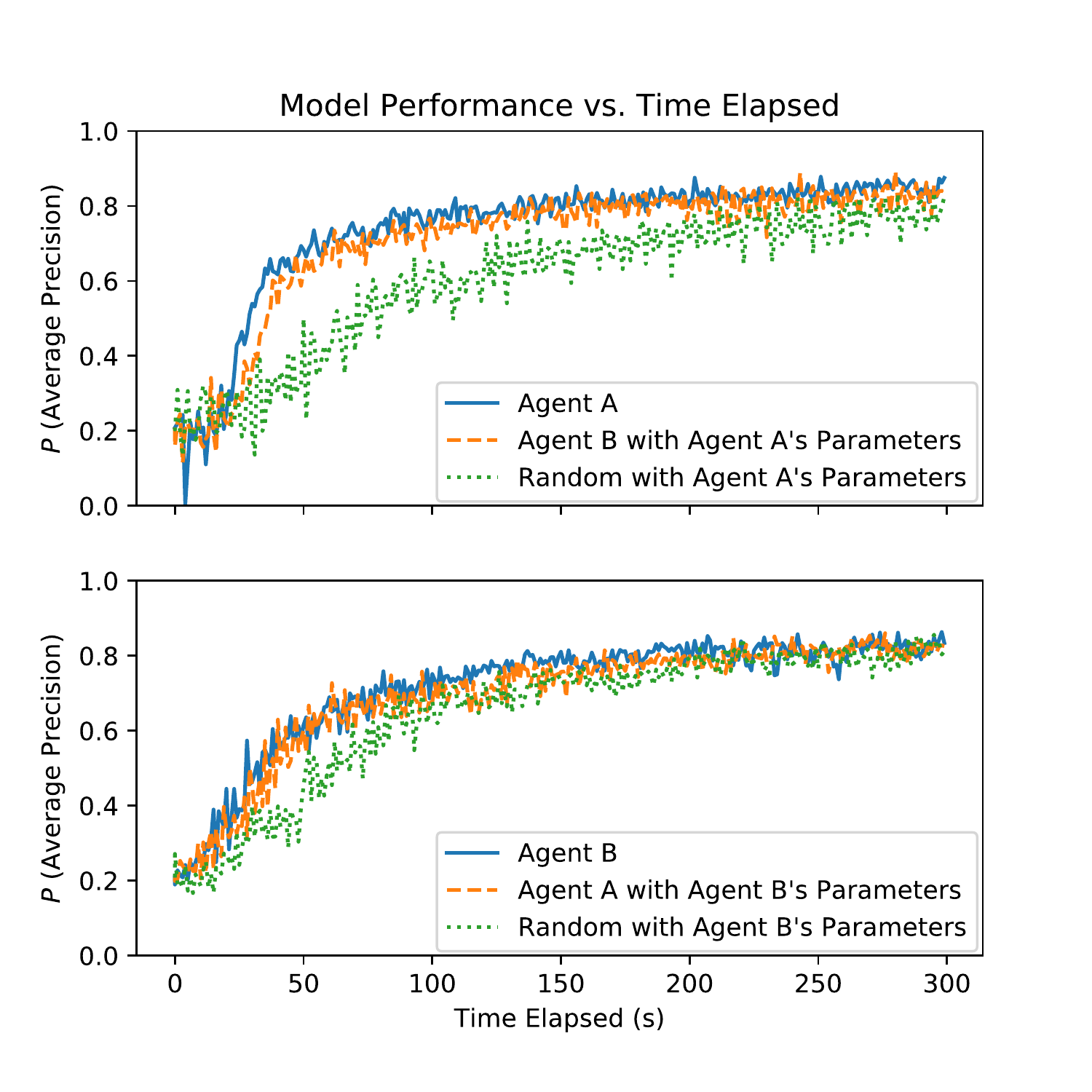}
    \caption{(top) Plot of $P_{t_{elapsed}}$ vs. elapsed time for 50 simulated scenarios, for a simulated platform with Agent A's parameters. (bottom) Plot of $P_{t_{elapsed}}$ vs. elapsed time for 50 simulated episodes, for a simulated platform with Agent B's parameters. The yellow dotted line on both represent mismatched agents. Performance is slightly worse on both simulations for mismatched agents.}
    \label{fig:sim_timeseries_300}
\end{figure}

Figure~\ref{fig:sim_timeseries_300} shows the increase in $P_{t_{elapsed}}$, for both agents. In the top plot, we simulate a drone platform with Agent A's parameters ($s_a = 2.5$ meters/second, $t_{train} = 0.305$ seconds), and in the bottom, we simulate a drone platform with Agent B's parameters ($s_a = 10.0$ meters/second, $t_{train} = 2.5$ seconds). The blue line in each plot represents the right agent for the right simulation, and the yellow dotted line is the wrong agent. A random policy is included for reference.

We do see a performance discrepancy between an agent trained for the particular simulated drone platform, and one that was not, with the bespoke agent consistently performing better than the other. This difference tapers off as more time elapses. Both scenes seem to have the same asymptote for average precision, and regardless of the policy, both agents approach that value as time passes. Another interesting trend is that both agents in the Agent A simulation climb faster in the first 50 seconds than both agents in the Agent B simulation. This makes sense---training rounds take far less time, and in the initial phase, movement is less critical.

\begin{figure}[ht] \centering
    \includegraphics[width=1.0\linewidth]{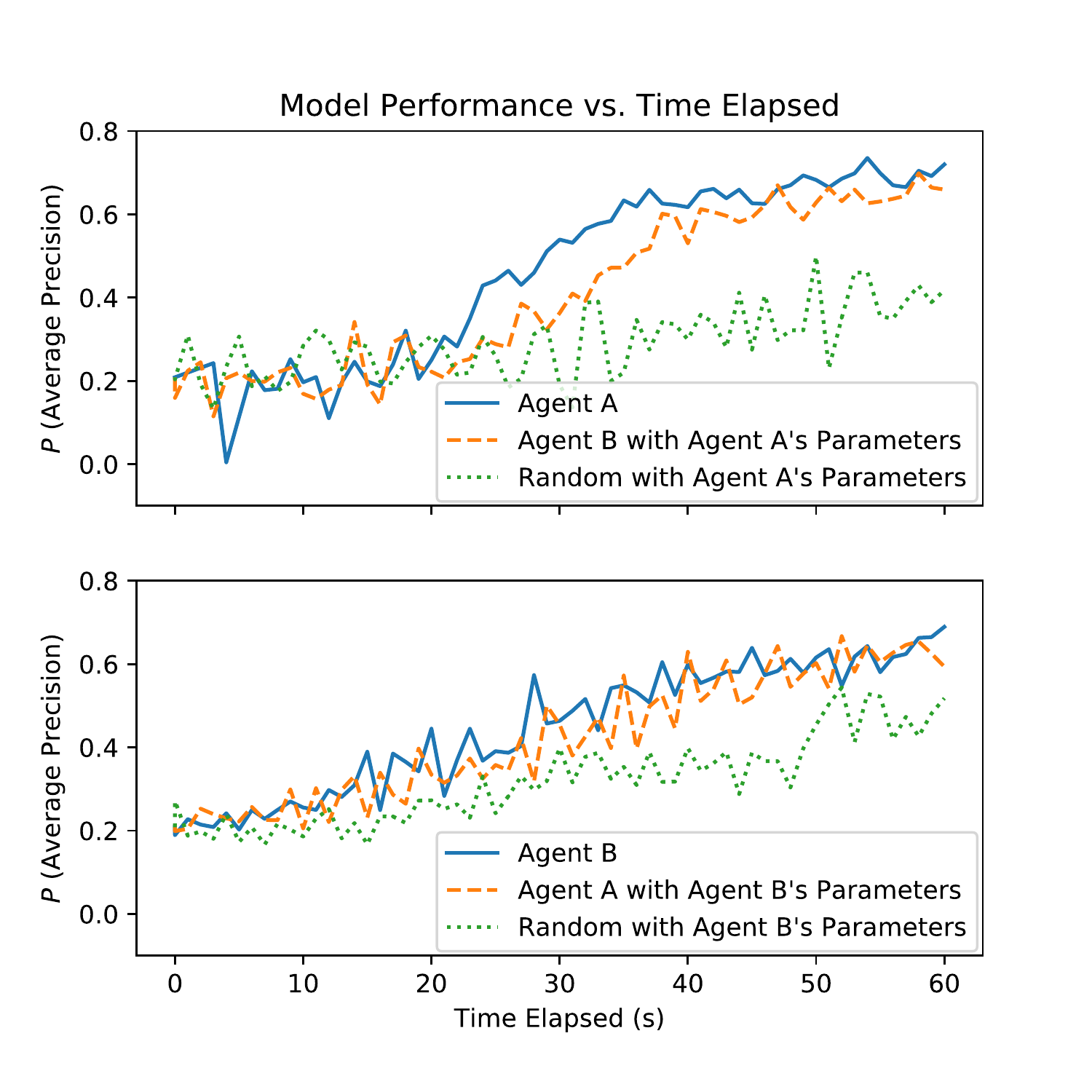}
    \caption{(top) Plot of $P_{t_{elapsed}}$ vs. elapsed time for the first minute of 50 simulated scenarios, for a simulated platform with Agent A's parameters. (bottom) Plot of $P_{t_{elapsed}}$ vs. elapsed time for the first minute of 50 simulated episodes, for a simulated platform with Agent B's parameters. The yellow dotted line on both represent mismatched agents. Performance is slightly worse on both simulations, with the difference being greater in the first one.}
    \label{fig:sim_timeseries_60}
\end{figure}

Figure~\ref{fig:sim_timeseries_60} shortens the episode to 60 seconds. We see that the difference between the two agents are more pronounced. We see the same trend reflected in Table~\ref{tab:agentavsb}. In particular, there is a gap around the 30 second mark between the two agents in the Agent A simulation, while no such gap exists in the second simulation. This is most likely from the fact that as $t_{train}$ increases, it begins to dominate the time elapsed for any movement step---regardless of where the drone is in the scene, the time for an action-step is most likely going to be equal to $t_{train}$.

\begin{table}[htb]
\centering
\resizebox{1.0\linewidth}{!}{%
  \begin{tabular}{ c | c | c | c | c }
    \hline
    & \multicolumn{2}{c|}{\scriptsize All 5 min} & \multicolumn{2}{c}{\scriptsize First 1 min} \\ \hline
    \scriptsize\textbf{Strategy} & $\overline{ITB}_{t_{elapsed}}$ & $\overline{ITB}_{u_{(1,f)}}$& $\overline{ITB}_{t_{elapsed}}$ & $\overline{ITB}_{u_{(1,f)}}$  \\ \hline
    Agent A &  \num{2.23e-3} &  \num{8.59e-3} &  \num{8.52e-3} &   \num{3.73e-2}  \\ \hline
    Agent B w/ A's params & \num{2.11e-3} &  \num{2.29e-2} &  \num{7.51e-3} &   \num{3.31e-2}   \\ \hline
    Random w/ A's params & \num{2.07e-3} &  \num{2.41e-2} &  \num{3.48e-3} &   \num{3.88e-2}   \\ \hline
    Agent B & \num{2.09e-3} &  \num{2.15e-2} &  \num{8.01e-3} &   \num{4.70e-2}   \\ \hline
    Agent A w/ B's params & \num{2.07e-3} &  \num{1.14e-2} &  \num{6.42e-3} &   \num{3.69e-2}    \\ \hline
    Random w/ B's params & \num{2.02e-3} &  \num{1.68e-2} &  \num{5.16e-3} &   \num{4.14e-2}    \\ \hline
  \end{tabular}}
\caption{Values for $\overline{ITB}_{u_{(1,f)}}$ and $\overline{ITB}_{t_{elapsed}}$ for Agents A and B.}
   \label{tab:agentavsb}
\end{table}

\begin{figure} \centering
    \includegraphics[width=1.0\linewidth]{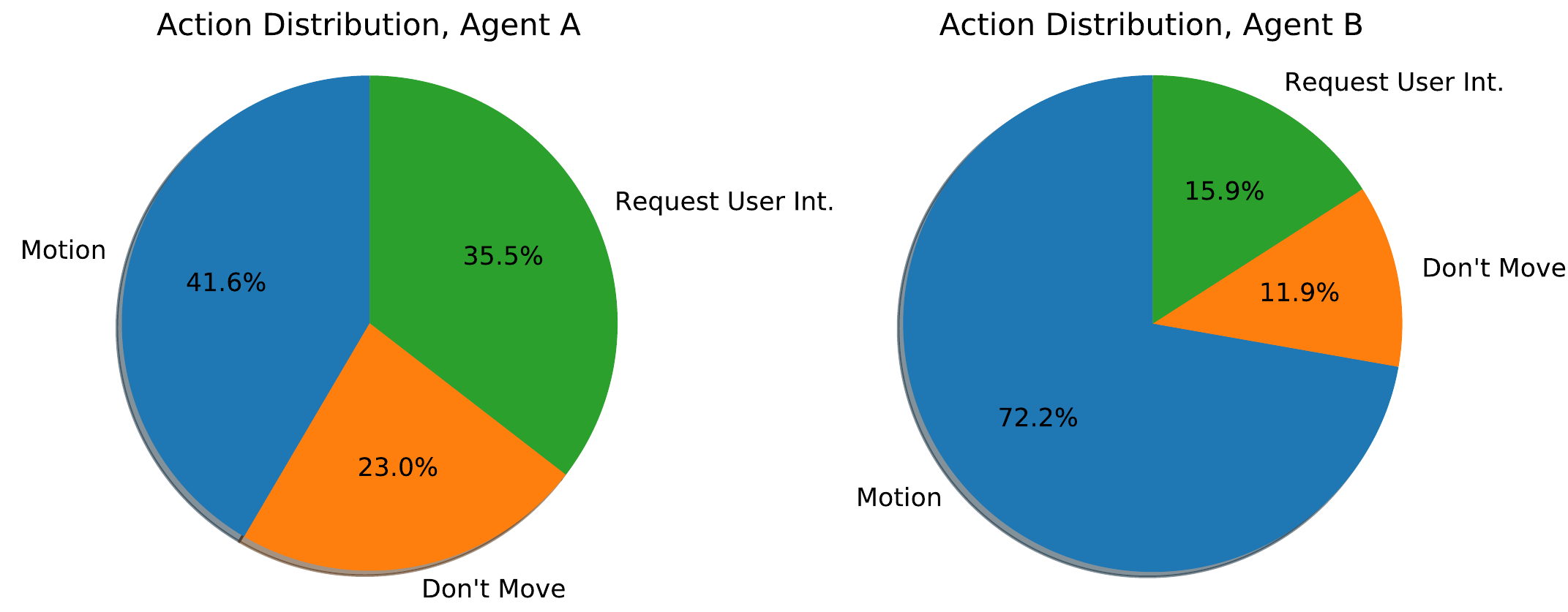}
    \caption{(left) Distribution of actions for Agent A, which has slow airspeed but fast training rounds. The agent learns to lean heavily on training rounds and asking the user rather than moving. (right) Distribution of actions for Agent B, which has fast airspeed but slow training rounds. The agent compensates for slow training by moving more, gathering more varied views more quickly.}
    \label{fig:curiosity_2d_actdistr}
\end{figure}

How does the behavior differ between agents trained with difference performance characteristics? Figure~\ref{fig:curiosity_2d_actdistr} show the distribution of actions for both agents during their 50 episodes. Note that the Motion category encompasses the forward, backwards, left, and right movements, and that Don't Move is not necessarily idle; the agent could be running training rounds. As expected, Agent A, trained for a slower drone but faster processing platform, uses more actions that require training rounds. Agent B compensates for its slow training rounds by moving around more. 


\subsection{$\overline{ITB}_{u_{(1,f)}}$ versus $\overline{ITB}_{t_{elapsed}}$}

\begin{figure} \centering
    \includegraphics[width=\linewidth]{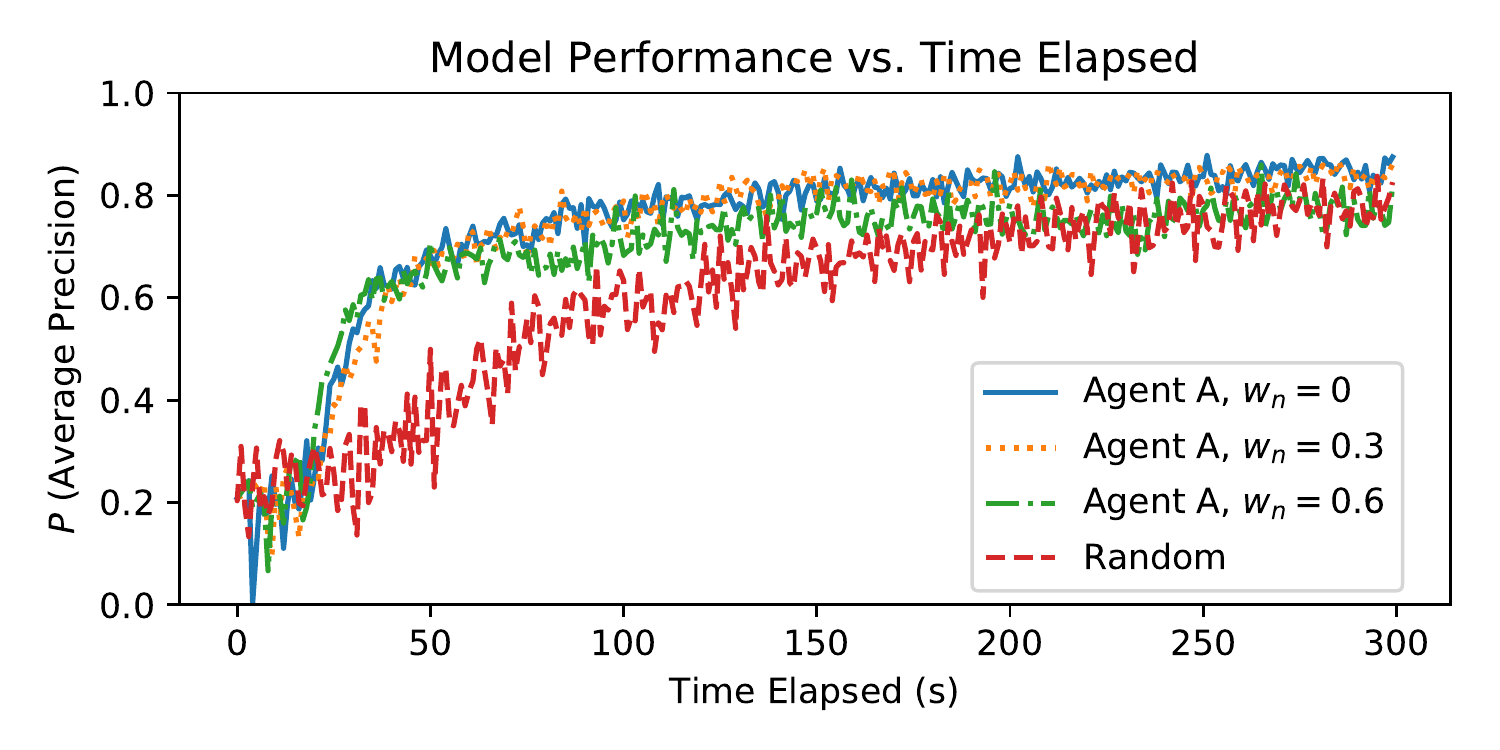}
    \caption{Plot of $P_{t_{elapsed}}$ for Agent A trained with $w_n = 0, 0.3,  0.6$. Random is shown as a reference. We see that all three agents produce the same curve shape (rising much faster than a random policy), but there is a tradeoff the performance of the model as we increase $w_n$. }
    \label{fig:change_wn}
\end{figure}

We noticed that our agents, while able to increase the $P_{t_{elapsed}}$ of the trainee much more rapidly than a random strategy, are not more efficient than random in terms of $\overline{ITB}_{u_{(1,f)}}$. In fact, Agent A performs substantially worse, asking the user for annotations much more frequently (Figure~\ref{fig:curiosity_2d_actdistr}). We then examined how the agent performs if we increase $w_n$, effectively increasing the negative impact of making user requests. To do this, we re-trained Agent A with increasing values of $w_n$. Figure~\ref{fig:change_wn} shows the plot of $P_{t_{elapsed}}$ for three values of $w_n = 0, 0.3, 0.6$, and Table~\ref{tab:wn_vary} records the values of $\overline{ITB}_{u_{(1,f)}}$ and $\overline{ITB}_{t_{elapsed}}$. 

\begin{table}[h]
\centering
\resizebox{\linewidth}{!}{%
  \begin{tabular}{ c | c | c | c | c }
    \hline
    & \multicolumn{2}{c|}{\scriptsize All 5 min} & \multicolumn{2}{c}{\scriptsize First 1 min} \\ \hline
    \scriptsize\textbf{Strategy} & $\overline{ITB}_{t_{elapsed}}$ & $\overline{ITB}_{u_{(1,f)}}$& $\overline{ITB}_{t_{elapsed}}$ & $\overline{ITB}_{u_{(1,f)}}$  \\ \hline
    Agent A, $w_n = 0$ &  \num{2.23e-3} &  \num{8.59e-3} &  \num{8.52e-3} &   \num{3.73e-2}  \\ \hline
    Agent A, $w_n = 0.3$ & \num{2.11e-3} &  \num{2.18e-2} &  \num{8.76e-3} &   \num{6.57e-2}   \\ \hline
    Agent A, $w_n = 0.6$ & \num{2.04e-3} &  \num{6.89e-2} &  \num{7.96e-3} &   \num{0.119}   \\ \hline
  \end{tabular}}
\caption{Values for $\overline{ITB}_{u_{(1,f)}}$ and $\overline{ITB}_{t_{elapsed}}$ of Agent A for varying values of $w_n$.}
   \label{tab:wn_vary}
\end{table}

We see in Table~\ref{tab:wn_vary} that the training benefit per user interaction $\overline{ITB}_{u_{(1,f)}}$ dramatically increases with $w_n$, with $w_n = 0.6$ producing more than 8 times more benefit per interaction than $w_n = 0$. However, there is a tradeoff to requesting less user interactions---we are less time-efficient ($\overline{ITB}_{u_{(1,f)}}$) by about 9\% for a 5-minute episode, and 7\% for the shortened 60-second episode. As seen in Figure~\ref{fig:change_wn}, all agents are still better than random, but the loss in model performance with increasing $w_n$ is significant. 

\begin{figure} \centering
    \includegraphics[width=\linewidth]{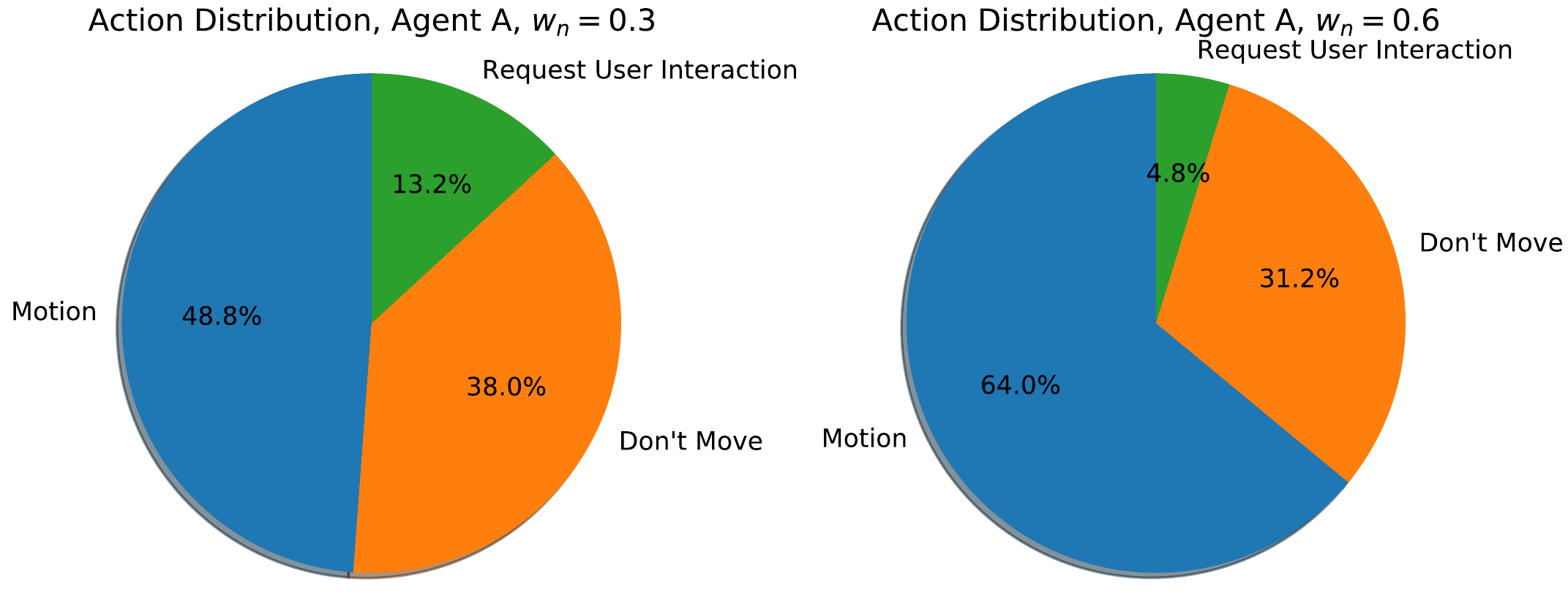}
    \caption{(left) Distribution of actions for Agent A with $w_n = 0.3$. (right) Distribution of actions for Agent A with $w_n = 0.6$. $w_n = 0$ is shown in Figure~\ref{fig:curiosity_2d_actdistr}. As $w_n$ increases, the agent prefers other actions besides asking the user for ground truth (user interactions).}
    \label{fig:curiosity_2d_actdistr_wn}
\end{figure}

In short, $w_n$ becomes a so-called ``annoyance parameter.'' By setting it higher, the user is asking the agent to ask for input less---at the sacrifice of the speed and effectiveness of learning. We can imagine using multi-objective reinforcement learning techniques~\cite{Tajmajer2017ModularValues,Mossalam2016Multi-ObjectiveLearning} to treat $w_n$ as an input to the agent during inference. This would enable the user to specify their own $w_n$ in the field---or have it adapt to current mission conditions.



\section{Conclusion}
\label{sec:conclusion}

In this paper, we examined how we can train a robotic agent to learn how to be curious, seeking out new data, and requesting human input when necessary, on its own within a two-dimension range-of-motion. Because it would be prohibitively difficult to perform deep reinforcement learning in reality, we train our agent within a simulated environment that trades realism for visual variety. Within this environment, we explored the design of a reward function that would incentivize the agent to learn how to learn quickly, taking into account the physical characteristics of the robotic system it is running on. We also showed that we can balance the agent's desire to learn quickly with its need for constant human input by weighing these elements in the reward function. We believe this approach is a step forwards in enabling robotic agents to be collaborative, rather than passive, students of their human operators, reducing overall operator mental load. 

\section{Acknowledgments}

We would like to thank Gerald Scott, Ashley Hobson, and the rest of the U.S. Naval Postgraduate School's Joint Interagency Field Experimentation (JIFX) staff for their continued support, without which these field studies would not have been possible. We would also like to thank Cef Ramirez for his extensive work on the infrastructure which makes real-time data collection possible. 



\section{Appendix}

\begin{algorithm}
\caption{Curiosity agent during an episode. At each time-step, the agent can either not move, move the drone, or request ground truth. In addition, if an object is being tracked by the trainee, a training event is initiated for the trainee. If updateTracker fails, $b_{track}$ is set to None.}\label{alg:curiosityagent2}
\begin{algorithmic}[1]
\State $\alpha \gets 0$ \Comment{Position in orbit}
\State $r \gets r_{max}$ \Comment{Distance from center of orbit}
\State $b_{track} \gets $None \Comment{Tracking BBox}
\State $v \gets $getCurrentView($p_{orbit}$)
\State $t_{elapsed} = 0$
\For{$i$ = (0,$n$)}
  \State $a_i \gets $chooseAction(A) \Comment{Agent chooses next action}
    \If{$a_i = 0$} \Comment{Don't move}
    	\If{$b_{track}$ is not None}
        \State trainingRound($v$, $b_{track}$)
        \EndIf
 	\ElsIf{$a_i = 1$} \Comment{Request User action}
    	\State $b_{track} \gets $requestGroundTruth($v$)
        \State trainingRound($v$, $b_{track}$)
    \ElsIf{$a_i = 2$} \Comment{Move left}
    	\If{$b_{track}$ is not None}
        \State trainingRound($v$, $b_{track}$)
        \EndIf
        $\alpha \gets \alpha + \Delta \alpha$ \Comment{Change orbit position}
    \ElsIf{$a_i = 3$} \Comment{Move right}
    	\If{$b_{track}$ is not None}
        \State trainingRound($v$, $b_{track}$)
        \EndIf
        $\alpha \gets \alpha - \Delta \alpha$ \Comment{Change orbit position}
        \ElsIf{$a_i = 4$} \Comment{Move forward}
          \If{$b_{track}$ is not None}
          \State trainingRound($v$, $b_{track}$)
          \EndIf
        \If{$r > r_{min}$}
        \State $r \gets r - \Delta r$ \Comment{Change orbit distance}
        \EndIf
        \ElsIf{$a_i = 5$} \Comment{Move backward}
          \If{$b_{track}$ is not None}
          \State trainingRound($v$, $b_{track}$)
          \EndIf
        \If{$r < r_{max}$}
        	\State $r \gets r + \Delta r$ \Comment{Change orbit distance}
        \EndIf
    \EndIf
    \State $v \gets $getCurrentView($r$, $\alpha$)
    \State $b_{track} \gets $updateTracker($v_{i+1}$)
     \State $t_i \gets $computeElapsedTime($r$, $\alpha$, $a_i$)
    \State $t_{elapsed} \gets t_{elapsed} + t_i$
\EndFor
\State \textbf{return}
\end{algorithmic}
\end{algorithm}

{\small
\bibliographystyle{aaai}
\bibliography{Mendeley}
}

\end{document}